\documentclass{bmvc2k}

\usepackage{times}
\usepackage{epsfig}
\usepackage{graphicx}
\usepackage{amsmath}
\usepackage{amssymb}
\usepackage{makecell}
\usepackage{caption}
\usepackage[lined,ruled,commentsnumbered]{algorithm2e}
\usepackage{bbding}
\usepackage{pifont}
\usepackage{wasysym}
\usepackage{multirow}
\usepackage{float}

\newcommand{\argmaxA}{\mathop{\mathrm{argmax}}\limits}

\title{Guided Zoom: Questioning Network Evidence for Fine-Grained Classification}

\addauthor{Sarah Adel Bargal$^*$}{sbargal@bu.edu}{1}{ }{ }
\addauthor{Andrea Zunino$^*$}{andrea.zunino@iit.it}{2}{ }{ }
\addauthor{Vitali Petsiuk}{vpetsiuk@bu.edu}{1}{ }{ }
\addauthor{Jianming Zhang}{jianmzha@adobe.com}{3}{ }{ }
\addauthor{Kate Saenko}{saenko@bu.edu}{1}{ }{ }
\addauthor{Vittorio Murino}{vittorio.murino@iit.it}{2}{$^,$}{$^\textrm{4}$}
\addauthor{Stan Sclaroff}{sclaroff@bu.edu}{1}{ }{ }

\addinstitution{
 Department of Computer Science \\ Boston University 
}
\addinstitution{
 Pattern Analysis \& Computer Vision \\ Istituto Italiano di Tecnologia 
}
\addinstitution{Adobe Research 
}
\addinstitution{Department of Computer Science \\ University of Verona 
}

\runninghead{BARGAL, ZUNINO, PETSIUK, ZHANG, SAENKO, MURINO, SCLAROFF}{GUIDED ZOOM}

\def\ie{\emph{i.e}\bmvaOneDot}

\def\etal{\emph{et al}\bmvaOneDot}

\begin{document}

\maketitle

\begin{abstract}
We propose {\fontfamily{qcr}\selectfont Guided Zoom}, an approach that utilizes spatial grounding of a model's decision to make more informed predictions. 
It does so by making sure 
the model has ``the right reasons'' for a prediction, 
defined as reasons that are coherent with those used to make similar correct decisions at training time. 
The reason/evidence upon which a deep convolutional neural network makes a prediction is defined to be the spatial grounding, in the pixel space, for a specific class conditional probability in the model output. {\fontfamily{qcr}\selectfont Guided Zoom} examines how reasonable such evidence is for each of the top-$k$ predicted classes, rather than solely trusting the top-1 prediction. We show that {\fontfamily{qcr}\selectfont Guided Zoom} improves the classification accuracy 
of a deep convolutional neural network model and obtains state-of-the-art results on three fine-grained classification benchmark datasets. 

\end{abstract}

\section{Introduction}

For state-of-the-art deep single-label classification models, the correct class is often in the top-$k$ predictions, leading to a top-$k$ $(k=2,3,4, \dots)$ accuracy that is significantly higher than the top-1 accuracy. This accuracy gap is more pronounced in fine-grained classification tasks, where the differences between classes are quite subtle \cite{khosla2011novel,CUB,maji2013fine,krause20133d}. For example, the Stanford Dogs fine-grained dataset on which we report results has a top-1 accuracy of 86.9\% and a top-5 accuracy of 98.9\%. Exploiting the information provided in the top-$k$ predicted classes can boost the final prediction of a model. 

In this work, we do not completely trust a Convolutional Neural Network (CNN) model's top-1 prediction as it does not solely depend on the visual evidence in the input image, but can depend on other artifacts such as dataset bias or unbalanced training data. Instead, we target answering the following question: \textit{is the evidence upon which the prediction is made reasonable?} Evidence is defined to be the grounding, in pixel space, for a specific class conditional probability in the model output. 
We propose {\fontfamily{qcr}\selectfont Guided Zoom}, which uses evidence grounding as the signal to localize discriminative image regions and to assesses how much one can trust a CNN prediction over another. 

Since fine-grained classification requires focusing on details, the localization of discriminative object parts is crucial. Supervised approaches utilize part bounding box annotations \cite{zhang2016spda,zhang2014part,huang2016part} or have humans in the loop to help reveal discriminative parts \cite{deng2013fine}. Our approach does not require part annotations, thus 
it is more scalable compared to supervised approaches. 
Weakly supervised attention based methods require training CNNs with attention module(s) \cite{fu2017look,mnih2014recurrent,sun2018multi,zheng2017learning}, while our approach is a generic framework able to improve the performance of any CNN model, even if it was not trained using an attention module. 




\begin{figure}[t]
\centering
\includegraphics[width=0.65\linewidth, trim={0cm 0cm 0cm 0cm},clip]{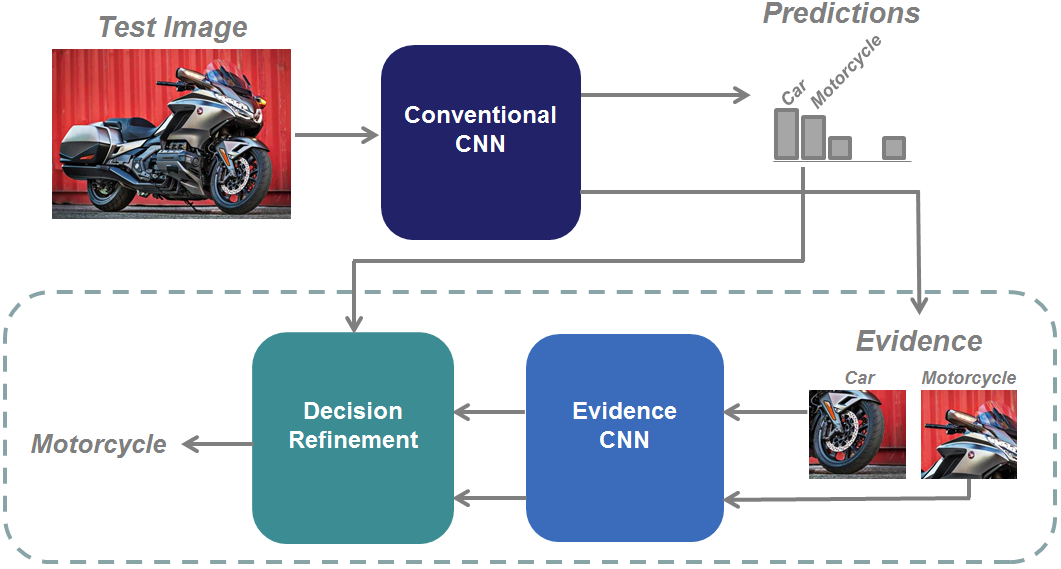}
\vspace{1.5em}
\caption[Pipeline of Guided Zoom]{Pipeline of {\fontfamily{qcr}\selectfont Guided Zoom}. A conventional CNN outputs class conditional probabilities for an input image. Salient patches could reveal that evidence is weak. 
We refine the class prediction of the conventional CNN by introducing two modules: 1) \textit{Evidence CNN} determines the consistency between the evidence of a test image prediction and that of correctly classified training examples of the same class. 2) \textit{Decision Refinement} uses the output of \textit{Evidence CNN} to refine the prediction of the conventional CNN.}
\label{figure:pipeline}
\end{figure}

\begin{figure*}[t]
\centering
\includegraphics[width=\linewidth, trim={0cm 0.1cm 0cm 0cm},clip]{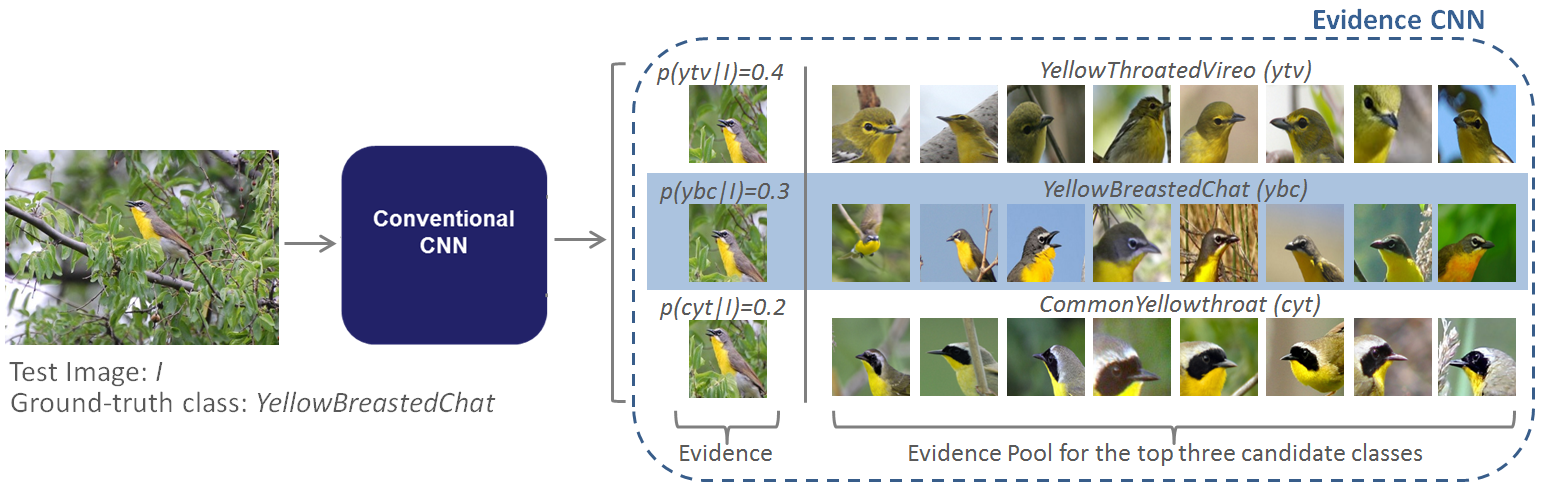}
\vspace{0.5em}
\caption[Consistency with pool evidence]{A conventional CNN is used to obtain salient image regions that highlight the evidence for predictions, together with the predicted class conditional probabilities. In our model, fine-grained classification decisions are improved by comparing consistency of the evidence for the incoming test image with the evidence seen for correct classifications in training. In this demonstration, although the conventional CNN predicts with highest probability the class \textit{YellowThroatedVireo}, \textit{Evidence CNN} is able to provide guidance for predicting the ground-truth class \textit{YellowBreastedChat} (highlighted in blue) due to visual similarity of the evidence of this class with that of the generated evidence pool.}
\label{figure:evidence_cnn}
\end{figure*}

{\fontfamily{qcr}\selectfont Guided Zoom} zooms into the evidence used to make a preliminary decision at test time and compares it with a reference pool of (evidence, prediction) pairs. The pool is constructed by accumulating (evidence, prediction) pairs for correctly classified training examples, providing additional free supervision from training data. As demonstrated in Fig.~\ref{figure:pipeline}, our approach examines whether or not the evidence used to make the prediction is coherent with training evidence of correctly classified images. This is performed 
by the \textit{Evidence CNN} module, which aids the \textit{Decision Refinement} module to come up with a refined prediction. The main goal of {\fontfamily{qcr}\selectfont Guided Zoom} is ensuring that evidence of the refined class prediction is more coherent with the training evidence of that class, compared to evidence of any of the other candidate top classes (see Fig.~\ref{figure:evidence_cnn}). By examining network evidence, we demonstrate improved accuracy and achieve state-of-the-art results on three fine-grained classification benchmark datasets. 





\section{Related Work}
\label{rel_work}

\hspace{1em} \textbf{Evidence Grounding.} The evidence behind a deep model's prediction for visual data results in highlighting the importance of image regions in making a prediction. Fong \etal \cite{8237633} exhaustively perturb image regions to find the most discriminative evidence. Petsiuk \etal \cite{Petsiuk2018rise} probe black-box CNN models with randomly masked instances of an image to find class-specific evidence. While such methods are bottom-up approaches, others are top-down and start from a high-level feature representation and then propagate down to the image level 
\cite{deconv,simonyan2013deep,grad,CAM,selvaraju2017grad,zhang2018top}. For example, Selvaraju \etal \cite{selvaraju2017grad} exploit a weighted sum of the last convolutional feature maps to obtain the class activation maps. Zhang \etal \cite{zhang2018top} use network weights and activations to obtain class activation maps, then highlight image cues unique to a specific class by eliminating evidence that also contributes to other classes. 

The result of evidence grounding is often referred to as saliency. Saliency is being widely used for many computer vision tasks including spatial semantic segmentation \cite{li2018tell,zhou2018weakly,wei2017object}, spatial object localization \cite{zhang2018adversarial,zhang2018top}, and temporal action localization \cite{bargal2017excitation}. Saliency has been less exploited for improving model classification. Cao \etal \cite{cao2015look} use weakly supervised saliency to feed highly salient regions into the same model that generated them to get more prediction probabilities for the same image and improve classification accuracy at test time. In contrast, we use weakly supervised saliency to examine whether the obtained evidence is coherent with the evidence used at training time for correctly classified examples. Zunino \etal \cite{zunino2018excitation} use spatial grounding at training time to improve model classification by dropping neurons corresponding to high-saliency patterns for regularization. In contrast, we propose an approach to improve model classification at test time. 

\textbf{Fine-grained classification.} The key module in fine-grained classification is finding discriminative parts. Some approaches use supervision to find such discriminative features, \ie use annotations for whole objects and/or for semantic parts. 
Zhang \etal \cite{zhang2014part} train part models such that the head/body can be compared; however, this requires a lot of annotation of parts. Krause \etal \cite{krause2015fine} use whole object annotations and no part annotations. Branson \etal \cite{branson2014bird} normalize the pose of object parts before computing a deep representation for them. Zhang \etal \cite{zhang2016spda} introduce part-abstraction layers in the deep classification model, enabling weight sharing between the two tasks. Huang \etal \cite{huang2016part} introduce a part-stacked CNN which encodes part and whole object cues in parallel based on supervised part localization. Wang \etal \cite{wang2016mining} retrieve neighboring images from the dataset, those having similar object pose, and automatically mine discriminative triplets of patches with geometric constraints as the image representation. Deng \etal \cite{deng2013fine} include humans in the loop to help select discriminative features. Subsequent work of Krause \etal \cite{krause2016unreasonable} does not use whole or part annotations, but augments fine-grained datasets by collecting web images and experimenting with filtered and unfiltered versions of them. Wang \etal \cite{wang2015multiple} use an ontology tree to obtain hierarchical multi-granularity labels. In contrast to such approaches, we do not require any object or part annotations at train or test time and do not use additional data or hierarchical labels.


Other approaches are weakly supervised. Such approaches only require an image label, and our approach lies in this category. Lin \etal \cite{lin2015bilinear} demonstrate the applicability of  a bilinear CNN model in the fine-grained classification task. Recasens \etal \cite{recasens2018learning} propose a distortion layer based on saliency to improve the input image sampling that demonstrates improvement for the fine-grained classification task. Sun \etal \cite{sun2018multi} implement an attention module that learns to localize different parts and a correlation module that enforces coherency in the correlations among different parts in training. Fu \etal \cite{fu2017look} learn where to focus by recurrently zooming into one location from coarse to fine using a recurrent attention CNN. In contrast, we are able to zoom into multiple image locations. Zhang \etal \cite{zhang2016picking} use convolutional filters as part detectors since the responses of distinctive filters usually focus on consistent parts. Zhao \etal \cite{zhao2017diversified} use a recurrent soft attention mechanism that focuses on different parts of the image at every time step. This work enforces a constraint to minimize the overlap of attention maps used in adjacent time steps to increase the diversity of part selection. Zheng \etal \cite{zheng2017learning} implement a multiple attention convolutional neural network with a fully-connected layer, combining the softmax for each part with one classification loss function. Cui \etal \cite{cui2017kernel} introduce a kernel pooling scheme and also demonstrate benefit to the fine-grained classification task. Jaderberg \etal \cite{jaderberg2015spatial} introduce spatial transformers for convolutional neural networks that
learn invariance to translation, scale, rotation and more generic warping, showing improvement for the task of fine-grained classification. 

In contrast, our approach assesses whether the network evidence used to make a prediction is reasonable, \ie if it is coherent with the evidence of correctly classified training examples of the same class. We use multiple salient regions, thus eliminating error propagation from incorrect initial saliency localization, and implicitly enforce part-label correlations enabling the model to make more informed predictions at test time.

\section{Guided Zoom}
\label{guided_zoom}

{\fontfamily{qcr}\selectfont Guided Zoom} uses the evidence of a prediction to improve classification performance by comparing the coherence of such evidence with a pool of ``reasonable'' class-specific evidence. We now describe how {\fontfamily{qcr}\selectfont Guided Zoom} utilizes multiple discriminative evidence, does not require part annotations, and implicitly enforces part correlations. This is done through the main modules depicted in Fig.~\ref{figure:pipeline}: \textit{Evidence CNN} and \textit{Decision Refinement}.


\textbf{Evidence CNN.} Conventional CNNs trained for image classification output class conditional probabilities upon which predictions are made. The class conditional probabilities are the result of some corresponding evidence in the input image. 
From correctly classified training examples, we generate a reference pool $\mathcal{P}$ of (evidence, prediction) pairs over which the \textit{Evidence CNN} will be trained for the same classification task. We recover/ground such evidence using a top-down spatial grounding technique: contrastive Excitation Backprop (\textit{c}EB) \cite{zhang2018top}. Starting with a prior probability distribution, \textit{c}EB passes top-down signals through excitatory connections (having non-negative weights) of a CNN. Recursively propagating the top-down signal layer by layer, \textit{c}EB computes class-specific discriminative saliency maps from any intermediate layer in a partial single backward pass. This is done by setting the prior distribution in correspondence with the correct class to produce a \textit{c}EB saliency map for it as described in the experiments section. We extract one 150x150 pixel patch from the original image around the resulting peak saliency. 
This patch highlights the most discriminative evidence. However, the next most discriminative patches may also be good additional evidence for differentiating fine-grained categories. 

\begin{figure}[t]
\centering
\includegraphics[width=0.11\linewidth,height=0.11\linewidth, trim={0cm 0cm 0cm 0cm},clip]{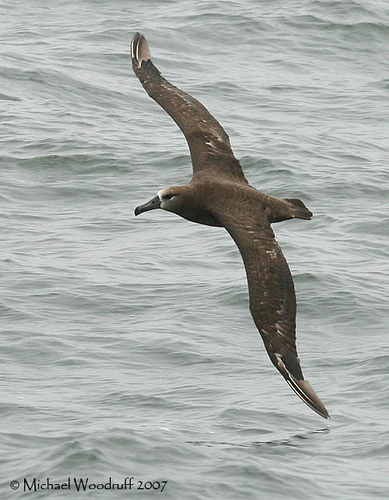}
\includegraphics[width=0.11\linewidth,height=0.11\linewidth, trim={0cm 0cm 0cm 0cm},clip]{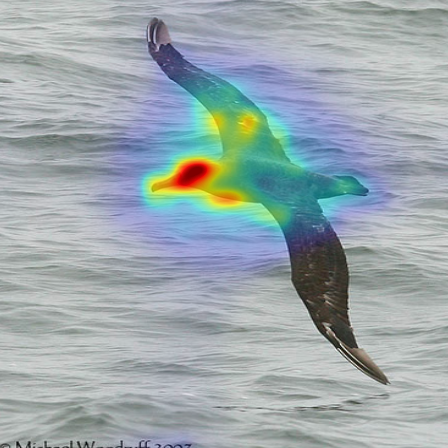}
\includegraphics[width=0.11\linewidth,height=0.11\linewidth, trim={0cm 0cm 0cm 0cm},clip]{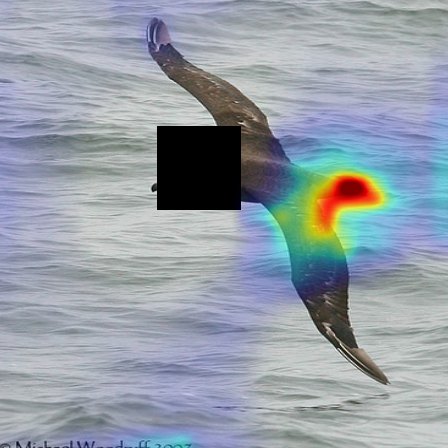}
\includegraphics[width=0.11\linewidth,height=0.11\linewidth, trim={0cm 0cm 0cm 0cm},clip]{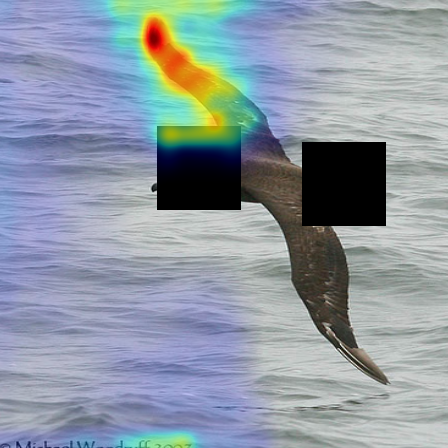}
\hspace{2.5em}
\includegraphics[width=0.11\linewidth, height=0.11\linewidth, trim={0cm 0cm 0cm 0cm},clip]{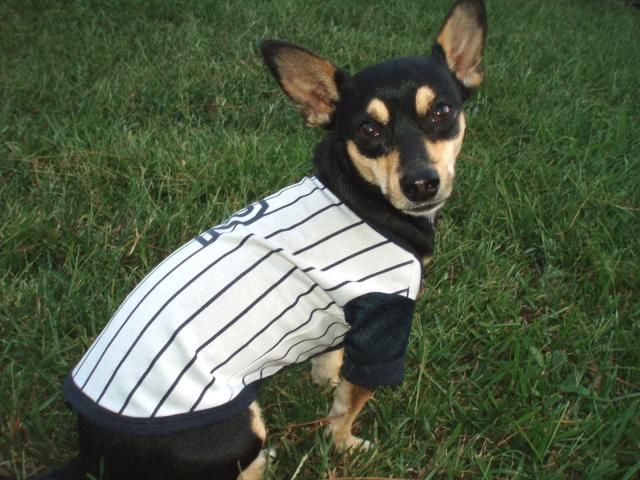}
\includegraphics[width=0.11\linewidth, trim={0cm 0cm 0cm 0cm},clip]{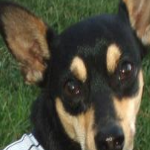}
\includegraphics[width=0.11\linewidth, trim={0cm 0cm 0cm 0cm},clip]{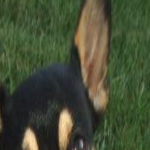}
\includegraphics[width=0.11\linewidth, trim={0cm 0cm 0cm 0cm},clip]{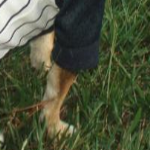}\\[0.2em]
\includegraphics[width=0.11\linewidth,height=0.11\linewidth, trim={0cm 0cm 0cm 0cm},clip]{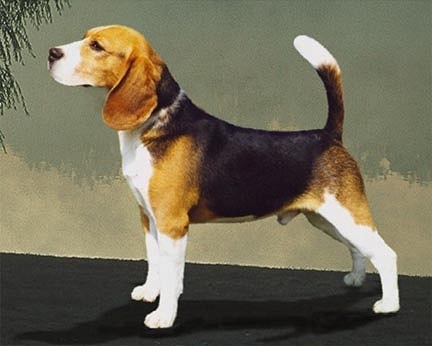}
\includegraphics[width=0.11\linewidth,height=0.11\linewidth, trim={0cm 0cm 0cm 0cm},clip]{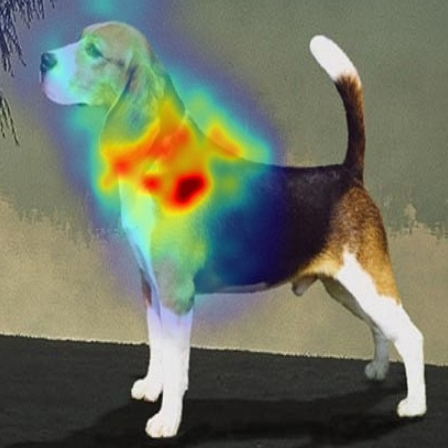}
\includegraphics[width=0.11\linewidth,height=0.11\linewidth, trim={0cm 0cm 0cm 0cm},clip]{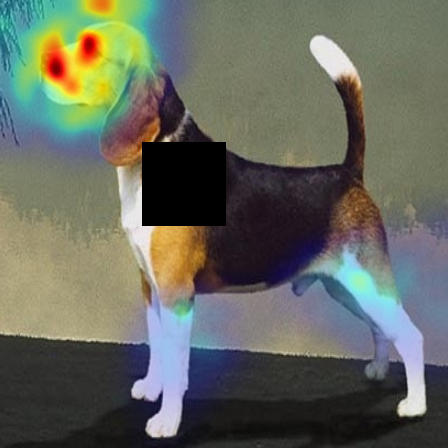}
\includegraphics[width=0.11\linewidth,height=0.11\linewidth, trim={0cm 0cm 0cm 0cm},clip]{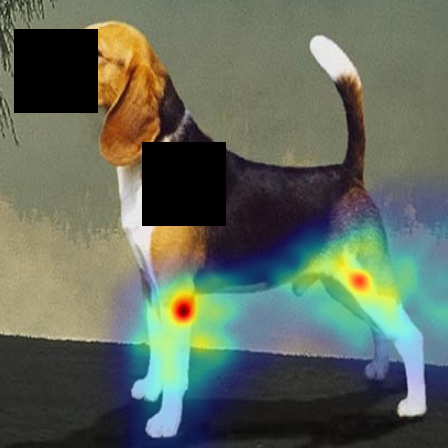}
\hspace{2.5em}
\includegraphics[width=0.11\linewidth, height=0.11\linewidth, trim={0cm 0cm 0cm 0cm},clip]{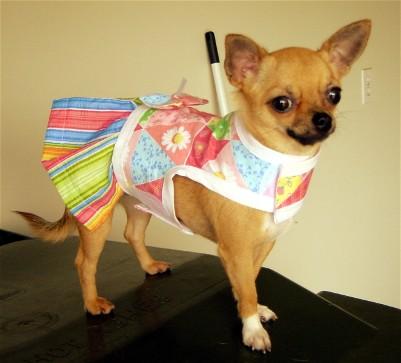}
\includegraphics[width=0.11\linewidth, trim={0cm 0cm 0cm 0cm},clip]{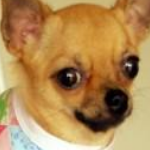}
\includegraphics[width=0.11\linewidth, trim={0cm 0cm 0cm 0cm},clip]{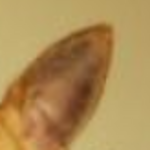}
\includegraphics[width=0.11\linewidth, trim={0cm 0cm 0cm 0cm},clip]{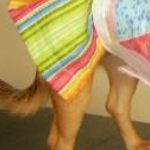}\\[0.2em]
\includegraphics[width=0.11\linewidth,height=0.11\linewidth, trim={0cm 0cm 0cm 0cm},clip]{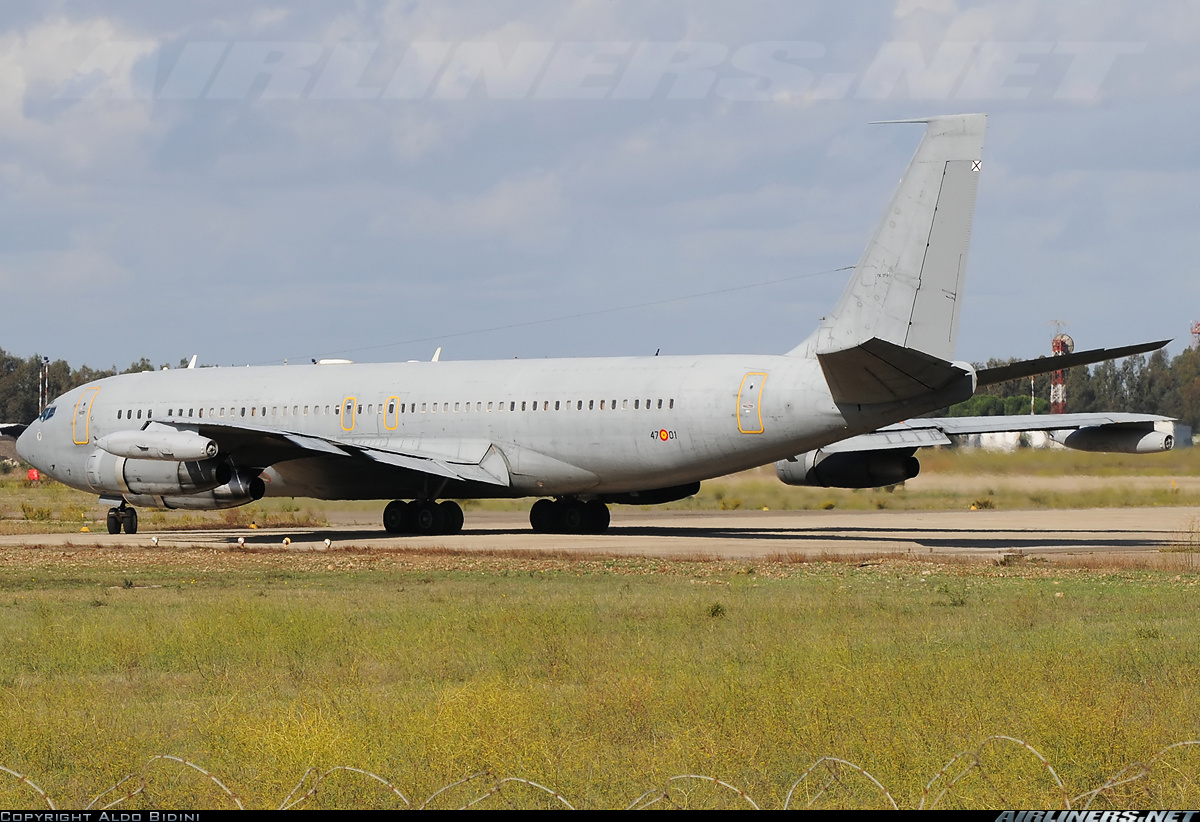}
\includegraphics[width=0.11\linewidth,height=0.11\linewidth, trim={0cm 0cm 0cm 0cm},clip]{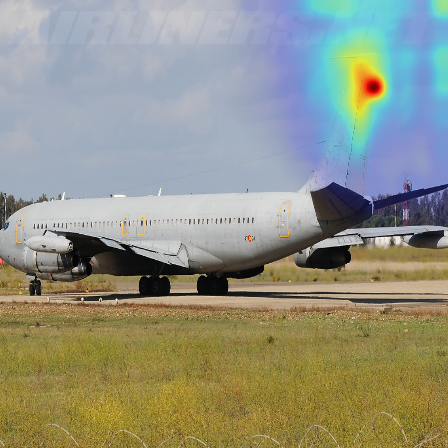}
\includegraphics[width=0.11\linewidth,height=0.11\linewidth, trim={0cm 0cm 0cm 0cm},clip]{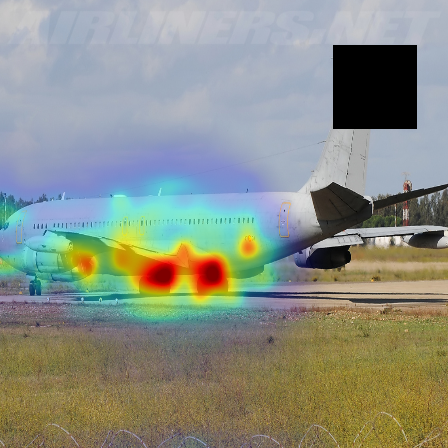}
\includegraphics[width=0.11\linewidth,height=0.11\linewidth, trim={0cm 0cm 0cm 0cm},clip]{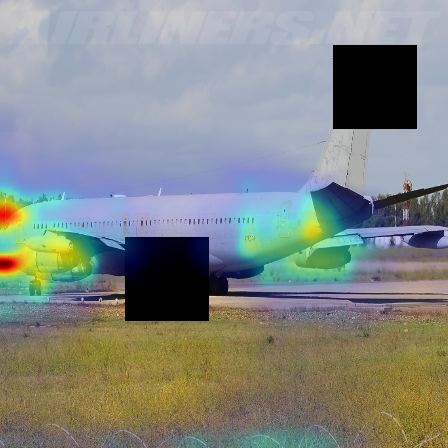}
\hspace{2.5em}
\includegraphics[width=0.11\linewidth, height=0.11\linewidth, trim={0cm 0cm 0cm 0cm},clip]{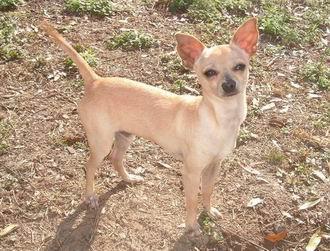}
\includegraphics[width=0.11\linewidth, trim={0cm 0cm 0cm 0cm},clip]{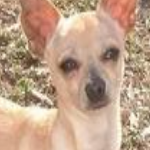}
\includegraphics[width=0.11\linewidth, trim={0cm 0cm 0cm 0cm},clip]{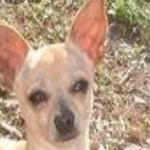}
\includegraphics[width=0.11\linewidth, trim={0cm 0cm 0cm 0cm},clip]{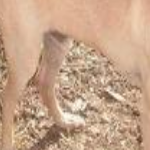}\\[0.2em]
\hspace*{-0.8em} original \hspace*{1.3em} $l=0$ \hspace*{1.8em} $l=1$ \hspace*{1.7em} $l=2$
\hspace*{4em} original \hspace*{1.2em} $l=0$ \hspace*{1.8em} $l=1$ \hspace*{1.6em} $l=2$
\vspace{1.75em}
\caption[Patch extraction using adversarial erasing]{\textbf{Adversarial Erasing.} Left: Sample image from each dataset to demonstrate the extraction of patches during two rounds of adversarial erasing: finding the first ($l=0$), second ($l=1$), and third ($l=2$) most-salient evidence. 
For example, the most salient evidence for the bird image is the head, followed by the tail, followed by the right wing. Right: Implicit part detection obtained as a result of two iterations of adversarial erasing for three images from the dog class \textit{Chihuahua}. Assigning the same class label to the different parts of a single dog image enforces implicit part-label correlation.}
\label{figure:adv_erasing2}
\end{figure}

Grounding techniques only highlight part(s) of an object. However, a more inclusive segmentation map can be extracted from the already trained model at test time using adversarial erasing. 
Inspired by the adversarial erasing work of Wei \etal \cite{wei2017object} in the domain of object segmentation, we extend its application to the domain of image classification by providing a data augmentation technique for network evidence. We augment our reference pool with patches resulting from performing an iterative adversarial erasing of the most discriminative evidence from the image. We notice that adversarial erasing results in implicit part localization from the most to least discriminative parts. Fig.~\ref{figure:adv_erasing2} (right) shows the patches extracted from two iterations of adversarial saliency erasing for sample images belonging to the class \textit{Chihuahua} from the Stanford Dogs Dataset. All patches (parts) extracted from this process inherit the ground-truth label of the original image. By labeling different parts with the same image ground-truth label, we are implicitly forcing part-label correlations in \textit{Evidence CNN}.

Including such additional evidence in our reference pool gives a richer description of the examined classes compared to models that recursively zoom into one location while ignoring other discriminative cues \cite{fu2017look}. We note that we add an evidence patch to the reference pool only if the removal of the previous salient patch does not affect the correct classification of the sample image. Erasing is performed by adding a black-filled 85x85 pixel square on the previous most salient evidence to encourage a highlight of the next salient evidence. This process is depicted in Fig.~\ref{figure:adv_erasing2} (left) for a sample bird species, dog species, and aircraft model.

Assuming $n$ training samples, for each sample $s^i$ where $i \in {1, \dots, n}$ we have $l+1$ evidence patches in the reference pool ${e_0^i, \dots, e_l^i}$. Patch $e_0^i$ is the most discriminative initial evidence, and ${e_1^i, \dots, e_l^i}$ is the set of the $l$ next discriminative evidence patches where $l\le L$ and $L$ is the number of adversarial erasing iterations performed ($L=2$ is used in our experiments). For example, $e_2^i$ is the third most-discriminative evidence, after the erasing of $e_0^i$ and $e_1^i$ from the original image. 
We then train a CNN model, the \textit{Evidence CNN}, on the classification of the generated evidence pool $\mathcal{P}$. 


\textbf{Decision Refinement.} At test time, we analyze whether the evidence upon which a prediction is made is reasonable. We do so by examining the consistency of a test (evidence, prediction) with our reference pool that is used to train \textit{Evidence CNN}. We exploit the visual evidence used for each of the top-$k$ predictions for
prediction refinement. The refined prediction will be inclined toward each of the top-$k$ classes by an amount proportional to how coherent its evidence is with the reference pool. For example, if the (evidence, prediction) of the second-top predicted class is more coherent with the reference pool of this class, then the refined prediction will be more inclined toward the second-top class.

Assuming test image $s^j$, where $j \in {1,\dots,m}$ and $m$ is the number of testing examples, $s^j$ is passed through the conventional CNN resulting in $v^{j,0}$, a vector of class conditional probabilities having some top-$k$ classes $c_1, \dots, c_k$ to be considered for the prediction refinement. We obtain the evidence for each of the top-$k$ predicted classes $e^{j,c_1}_0, \dots, e^{j,c_k}_0$, and pass each one through the \textit{Evidence CNN} to get the output class conditional probability vectors $v^{j,c_1}_0, \dots, v^{j,c_k}_0$. We then perform adversarial erasing to get the next most salient evidence $e^{j,c_1}_l, \dots, e^{j,c_k}_l$ and their corresponding class conditional probability vectors $v^{j,c_1}_l, \dots, v^{j,c_k}_l$, for $l \in {1, \dots, L}$. Finally, we compute a weighted combination of all class conditional probability vectors proportional to their saliency (a lower $l$ has more discriminative evidence and is therefore assigned a higher weight $w_l$). 
The estimated, refined class $c^j_{ref}$ is determined as the class having the maximum aggregate prediction in the weighted combination. Algorithm~\ref{alg2zoom} presents the steps used for decision refinement.

\begin{algorithm}[h!]
	\caption{Decision Refinement}\label{alg2zoom}    
	\vspace{.2 cm}
	\KwIn{$s^j, j \in {1, \dots, m}$ testing images, pre-trained conventional CNN, pre-trained \textit{Evidence CNN}, Grounding Method (GM), weights $w, w_0, \dots, w_L$}
    \vspace{.2 cm}
	\KwOut{Refined class for $s^j$: $c^j_{ref}$}
    \vspace{.2 cm} 		 
	{\bf Procedure}:\vspace{.2 cm}\\
    \nl For every test example $s^j, j \in 1, \dots, m$\\ \vspace{.2 cm}
	\nl \hspace*{1em} $v^{j,0}$:= conventional CNN($s^j$)\\ \vspace{.2 cm}
    \nl \hspace*{1em} $tot^j$:= $w * v^{j,0}$\\ \vspace{.2 cm}
	\nl \hspace*{1em} For $t \in {c_1, \dots, c_k}$, the top-$k$ classes of $v^{j,0}$ \\ \vspace{.2 cm}
        \nl \hspace*{2em} $e^{j,t}_0$:= GM($s^j$) w.r.t. class $t$\\ \vspace{.2 cm}
	\nl \hspace*{2em} $v^{j,t}_0$:= \textit{Evidence CNN}($e^{j,t}_0$) \\ \vspace{.2 cm}
     \nl \hspace*{2em}  $tot^j[t]:=tot^j[t] + w_0 * v^{j,t}_0[t]  $ \\	\vspace{.2 cm}
	 \nl \hspace*{2em} For $l \in 1, \dots, L$\\ \vspace{.2 cm}
    \nl \hspace*{3em} Adversarially erase $e^{j,t}_{l-1}$ from $s^j$ \\ \vspace{.2em}
    \nl \hspace*{3em} $e^{j,t}_l$:= GM($s^j$) w.r.t. class $t$\\ \vspace{.2 cm}
	\nl \hspace*{3em} $v^{j,t}_l$:= \textit{Evidence CNN}($e^{j,t}_l$) \\ \vspace{.2 cm}
    \nl \hspace*{3em}  $tot^j[t]:=tot^j[t] + w_l * v^{j,t}_l[t]  $ \\	\vspace{.2 cm}
    \nl \hspace*{1em} $c^j_{ref}:= \argmaxA_{c_1:c_k}(tot^j$)
 \end{algorithm}

\section{Experiments}
\label{exps}

In this section, we first present the fine-grained benchmark datasets we use to evaluate {\fontfamily{qcr}\selectfont Guided Zoom}. 
We then present the architecture and setup of our experiments, followed by a discussion of our experimental results. We note that although the datasets provide part annotations, we only use image-level class labels.

\vspace*{0.5em}
\textbf{Datasets.} We report experimental results on three fine-grained classification benchmark datasets following \cite{sun2018multi,fu2017look,zhao2017diversified,cui2017kernel,zheng2017learning}. 
\begin{itemize}
    \item CaltechUCSD (CUB-200-2011) Birds Dataset \cite{CUB} is a fine-grained dataset of 200 bird species consisting of $\sim$12K annotated images, split into $\sim$6K training images and $\sim$6K testing images. 
    \item Stanford Dogs Dataset \cite{khosla2011novel} is a fine-grained dataset of 120 dog species. This dataset includes $\sim$20K annotated images split into $\sim$12K and $\sim$8.5K images for training and testing respectively. 
    \item FGVC-Aircraft \cite{maji2013fine} is a fine-grained dataset of 100 different aircraft variants consisting of 10K annotated images, split into $\sim$7K training images and $\sim$3K testing images.
\end{itemize}

\textbf{Architecture and Setup.} To validate the benefit of {\fontfamily{qcr}\selectfont Guided Zoom}, we first purposely use a simple CNN baseline with a vanilla training scheme. We use a ResNet-101 \cite{He_2016_CVPR} 
network as the conventional CNN and baseline, extending the input size from the default 224x224 pixel to 448x448 pixel following \cite{sun2018multi,fu2017look,krause2016unreasonable}. The 448x448 pixel input image is a random crop from a 475x475 pixel input image at training time, and a center crop from a 475x475 pixel input image at test time. The reference pool is generated by extracting evidence patches from training images using a \textit{c}EB saliency map computed at the \textit{res4a}. We also then use the more complex MA-CNN architecture \cite{zheng2017learning} to demonstrate the generic nature of {\fontfamily{qcr}\selectfont Guided Zoom}. This follows the same input image sizes used for ResNet-101, and compute \textit{c}EB saliency maps at the \textit{conv3\_1} layer.

For the \textit{Evidence CNN}, we use a ResNet-101 architecture, but use the standard 224x224 pixel input size to keep the patches close to their original image resolution. This is a random crop from a 256x256 pixel input image at training time, and a center crop from a 256x256 pixel input image at test time. For both the conventional and \textit{Evidence} CNNs, and for all the three datasets, we use stochastic gradient descent, a batch size of 64, a starting learning rate of 0.001, multiplied by 0.1 every 10K iterations for 30K iterations, and momentum of 0.9. 

We demonstrate the benefit of using evidence information from the top-3 and top-5 predicted classes, so we set $k=3,5$ in our experiments. We perform two rounds of adversarial erasing in testing; we set $L=2$, and $w=0.4, w_0 = 0.3, w_1=0.2$, and $w_2=0.1$ giving lower weights to less discriminative evidence.

\begin{table*}[t]
\centering
\begin{tabular}{ll|ccc} 
\hline
& \textbf{Method} & \makecell{\textbf{Part / Whole}\\ \textbf{Annotation}} & \makecell{\textbf{Multiple}\\\textbf{Attention}} & \makecell{\textbf{Top-1} \\ \textbf{Accuracy (\%)}} \\ 
\hline
& DVAN \cite{zhao2017diversified} & x & \checkmark & 79.0 \\
& PA-CNN \cite{krause2015fine} & \checkmark & \checkmark & 82.8 \\
& MG-CNN \cite{wang2015multiple} & \checkmark & \checkmark & 83.0 \\
& B-CNN \cite{lin2015bilinear} & x & x & 84.1 \\
& RA-CNN \cite{fu2017look} & x & x & 85.3  \\
& PN-CNN \cite{branson2014bird} & \checkmark & \checkmark & 85.4  \\
& OSME + MAMC \cite{sun2018multi} & x  & \checkmark  & 86.5 \\
& MA-CNN \cite{zheng2017learning} & x &  \checkmark  & 86.5\\
\hline
\multirow{4}{*}{\rotatebox[origin=c]{90}{\textbf{\textit{Ours}}\hspace*{1.5em}}} & ResNet-101 & x & x & 82.3 \\& {\fontfamily{qcr}\selectfont Guided Zoom} (ResNet-101, $k$=3) & x & \checkmark  & 85.0 \\
& {\fontfamily{qcr}\selectfont Guided Zoom} (ResNet-101, $k$=5) & x  & \checkmark  & 85.4 \\
& {\fontfamily{qcr}\selectfont Guided Zoom} (MA-CNN, $k$=3)  & x  & \checkmark  & 87.6 \\
& {\fontfamily{qcr}\selectfont Guided Zoom} (MA-CNN, $k$=5)  & x  & \checkmark  & \bf{87.7} \\
\hline
\end{tabular}
\vspace{0.6em}
\caption[Fine-grained classification accuracy for the CUB-200-2011 Birds Dataset]{\textbf{CUB-200-2011 Birds Dataset.} We compare our classification accuracy with state-of-the-art weakly-supervised methods (do not use any sort of annotation apart from the image label) and some representative methods that use additional supervision such as part annotations for fine-grained classification of this dataset. We indicate which methods use multiple parts, and which focus on a single part using the multiple attention flag. Using part annotations implicitly entails multiple attention. We present results for our approach for $k$=3,5 using the top 3 (or 5) candidate classes to refine the final prediction.}
\label{table:Results_Birds}
\end{table*}

\begin{table*}[t]
\centering
\begin{tabular}{ll|ccc} 
\hline
& \textbf{Method} & \makecell{\textbf{Part / Whole}\\ \textbf{Annotation}} & \makecell{\textbf{Multiple}\\\textbf{Attention}} & \makecell{\textbf{Top-1} \\ \textbf{Accuracy (\%)}} \\ 
\hline
& DVAN \cite{zhao2017diversified} & x & \checkmark & 81.5 \\
& OSME + MAMC \cite{sun2018multi} & x  & \checkmark  & 85.2 \\
& RA-CNN \cite{fu2017look} & x & x & 87.3  \\
\hline
\multirow{4}{*}{\rotatebox[origin=c]{90}{\textbf{\textit{Ours}}\hspace*{-1em}}}& ResNet-101 & x & x & 86.9  \\ & {\fontfamily{qcr}\selectfont Guided Zoom} (ResNet-101, $k$=3) & x & \checkmark  &  88.4 \\
& {\fontfamily{qcr}\selectfont Guided Zoom} (ResNet-101, $k$=5) & x  & \checkmark  & \bf{88.5} \\
\hline
\end{tabular}
\vspace{0.6em}
\caption[Fine-grained classification accuracy for the Stanford Dogs Dataset]{\normalsize{\textbf{Stanford Dogs Dataset.} We compare our classification accuracy with state-of-the-art weakly-supervised methods (do not use any sort of annotation apart from the image label). We indicate which methods use multiple parts, and which focus on a single part using the multiple attention flag. Using part annotations implicitly entails multiple attention. We present results for our approach for $k$=3,5 using the top 3 (or 5) candidate classes to refine the final prediction.}
}
\label{table:Results_Dogs}
\end{table*}


\begin{table*}[t]
\vspace{-1.5em}
\centering
\begin{tabular}{ll|ccc} 
\hline
& \textbf{Method} & \makecell{\textbf{Part / Whole}\\ \textbf{Annotation}} & \makecell{\textbf{Multiple}\\\textbf{Attention}} & \makecell{\textbf{Top-1} \\ \textbf{Accuracy (\%)}} \\ 
\hline
& B-CNN \cite{lin2015bilinear} & x & x & 84.1 \\
& MG-CNN \cite{wang2015multiple} & \checkmark & \checkmark &  86.6 \\
& RA-CNN \cite{fu2017look} & x & x & 88.2  \\
& MDTP \cite{wang2016mining} & \checkmark & \checkmark & 88.4 \\
& MA-CNN \cite{zheng2017learning} & x  & \checkmark  & 89.9 \\
\hline
\multirow{4}{*}{\rotatebox[origin=c]{90}{\textbf{\textit{Ours}}\hspace*{1.5em}}} & ResNet-101 & x & x & 87.5 \\ & {\fontfamily{qcr}\selectfont Guided Zoom} (ResNet-101, $k$=3) & x & \checkmark  &  89.1\\
& {\fontfamily{qcr}\selectfont Guided Zoom} (ResNet-101, $k$=5) & x  & \checkmark  &  89.1\\
& {\fontfamily{qcr}\selectfont Guided Zoom} (MA-CNN, $k$=3) & x  & \checkmark  & \bf{90.7} \\
& {\fontfamily{qcr}\selectfont Guided Zoom} (MA-CNN, $k$=5) & x  & \checkmark  & \bf{90.7} \\
\hline
\end{tabular}
\vspace{0.6em}
\caption[Fine-grained classification accuracy for the FGVC-Aircraft Dataset]{\normalsize{\textbf{FGVC-Aircraft Dataset.} We compare our classification accuracy with state-of-the-art weakly-supervised methods (do not use any sort of annotation apart from the image label) and some representative methods that use additional supervision such as part annotations for fine-grained classification of this dataset. We indicate which methods use multiple parts, and which focus on a single part using the multiple attention flag. Using part annotations implicitly entails multiple attention. We present results for our approach for $k$=3,5 using the top 3 (or 5) candidate classes to refine the final prediction.}}
\label{table:Results_Aircrafts}
\vspace{-1em}
\end{table*}


\textbf{Results.} In this section, we demonstrate how training our \textit{Evidence CNN} module benefits from implicit part detection, a direct consequence of adversarial erasing that extracts the next most-salient evidence. 
Our framework targets providing complementary zooming on salient parts and use them for decision refinement. 

Table~\ref{table:Results_Birds} presents the results for the  CUB-200-2011 Birds dataset. Utilizing the top-3 class predictions together with their associated evidence, {\fontfamily{qcr}\selectfont Guided Zoom} boosts the top-1 accuracy from 82.3\% to 85.0\%. Utilizing the top-5 class predictions together with their associated evidence, {\fontfamily{qcr}\selectfont Guided Zoom} boosts the top-1 accuracy from 82.3\% to 85.4\%. Being a generic framework, we implement {\fontfamily{qcr}\selectfont Guided Zoom} to the image stream of the multi-zoom approach MA-CNN, obtaining state-of-the-art results on the CUB-200-2011 dataset. 

Table~\ref{table:Results_Dogs} presents the results for the  Stanford Dogs dataset. Utilizing the top-3 class predictions together with their associated evidence, {\fontfamily{qcr}\selectfont Guided Zoom} boosts the top-1 accuracy from 86.9\% to 88.4\%. 
Utilizing the top-5 class predictions together with their associated evidence, {\fontfamily{qcr}\selectfont Guided Zoom} boosts the top-1 accuracy from 86.9\% to 88.5\%. We achieve state-of-the-art results on the Stanford Dogs dataset.

Table~\ref{table:Results_Aircrafts} presents the results for the  FGVC-Aircraft dataset. Utilizing the top-3 class predictions together with their associated evidence, {\fontfamily{qcr}\selectfont Guided Zoom} boosts the top-1 accuracy from 87.5\% to 89.1\%. 
Utilizing the top-5 class predictions together with their associated evidence, {\fontfamily{qcr}\selectfont Guided Zoom} boosts the top-1 accuracy from 87.5\% to 89.0\%. 
{\fontfamily{qcr}\selectfont Guided Zoom} boosts the top-1 classification accuracy of MA-CNN obtaining state-of-the-art results on the FGVC-Aircraft dataset. 

{\fontfamily{qcr}\selectfont Guided Zoom} outperforms RA-CNN on all three datasets. From this we can conclude that our multi-zooming is more beneficial than a single recursive zoom. In addition, we demonstrate that {\fontfamily{qcr}\selectfont Guided Zoom} further improves the performance of the multi-zoom approach MA-CNN. Being a generic framework, {\fontfamily{qcr}\selectfont Guided Zoom} could be applied to any deep convolutional model for decision refinement within the top-$k$ predictions.



\section*{Conclusion}
In this work, we propose {\fontfamily{qcr}\selectfont Guided Zoom}, which utilizes explicit spatial grounding to refine a model's prediction at test time.
Our refinement module selects one of the top-$k$ model predictions having the most reasonable (evidence, prediction) pair, where ``most reasonable" is defined as the most consistent with respect to a pre-defined pool of training (evidence, prediction) pairs. The pool is populated by iteratively grounding and adversarially erasing the evidence for a correct prediction made by a conventional CNN.
{\fontfamily{qcr}\selectfont Guided Zoom} achieves state-of-the-art results on three fine-grained classification benchmark datasets.

\section*{Acknowledgments}
This work was supported in part by Defense Advanced Research Projects Agency (DARPA) Explainable Artificial Intelligence (XAI) program and Intelligence Advanced Research Projects Activity (IARPA) via Department of Interior/ Interior Business Center (DOI/IBC) contract number D17PC00341. The U.S. Government is authorized to reproduce and distribute reprints for Governmental purposes notwithstanding any copyright annotation thereon. Disclaimer: The views and conclusions contained herein are those of the authors and should not be interpreted as necessarily representing the official policies or endorsements, either expressed or implied, of DARPA, IARPA, DOI/IBC, or the U.S. Government.

\bibliography{egbib}

\end{document}